\documentclass[sigconf]{acmart}

\usepackage{multirow}
\usepackage{bm}
\usepackage{bbding}
\usepackage{subfig}
\usepackage{pgfplots}
\pgfplotsset{width=10cm,compat=1.16}
\usepackage{algorithm}
\usepackage{algorithmic}

\copyrightyear{2026}
\acmYear{2026}
\setcopyright{cc}
\setcctype{by}
\acmConference[ICCAD '26]{IEEE/ACM International Conference on Computer-Aided Design}{November 08--12, 2026}{San Jose, CA, USA}
\acmBooktitle{IEEE/ACM International Conference on Computer-Aided Design (ICCAD '26), November 08--12, 2026, San Jose, CA, USA}
\acmDOI{10.1145/3831252.3834128}
\acmISBN{979-8-4007-2873-0/2026/11}

\begin{document}


\title{MatMoE: Matryoshka Mixture-of-Experts with Dynamic Mixed-Precision Quantization for Efficient Inference}

\author{%
Renjie Wei\textsuperscript{1,2},
Haochen Huang\textsuperscript{1},
Chuyu Qiu\textsuperscript{4,1},
Jinqi Wen\textsuperscript{1},
Dehao Xu\textsuperscript{5,1},
Meng Li\textsuperscript{1,2,3 \(\dagger\)}
}

\affiliation{%
\institution{%
\textsuperscript{1}Institute for Artificial Intelligence
\&
\textsuperscript{2}School of Integrated Circuits,
Peking University, Beijing, China
\\
\textsuperscript{3}Beijing Advanced Innovation Center for Integrated Circuits, Beijing, China \\
\textsuperscript{4}Sun Yat-sen University,
Guangzhou, China
\quad
\textsuperscript{5}Southeast University,
Nanjing, China
}
\city{}
\country{}
}

\email{} 

\renewcommand{\shortauthors}{Wei et al.}

\thanks{%
\textsuperscript{\(\dagger\)}Corresponding author: meng.li@pku.edu.cn.
}

\definecolor{color1}{RGB}{80,125,206}
\definecolor{color2}{RGB}{255,194,78}

\begin{abstract}


Mixture-of-Experts (MoE) architecture has been widely adopted in large language models (LLMs) since it reduces the computational cost through sparse expert activation. 
However, the massive parameter scale of MoE models hinders their deployment on resource-constrained edge devices.
Although existing mixed-precision quantization reduces the model size, they are inherently static, fixing the bit-width of each expert across decoding steps. 
Such static schemes yield suboptimal performance as they overlook the dynamic variation in expert importance during decoding. 
Furthermore, they fail to adapt to the dynamic latency constraint 
driven by the fluctuating system utilization.
To address these challenges, we propose MatMoE, a dynamic mixed-precision quantization framework for efficient MoE inference. 
MatMoE features: (1) a Matryoshka MoE PTQ algorithm that enables experts to operate flexibly at multiple precisions within a stringent memory budget; (2) a floor-and-compensation quantization method that eliminates redundant memory accesses in the Matryoshka representation; and (3) a dynamic bit-width allocation strategy that determines the precision for each expert based on their routing scores to satisfy the dynamic latency requirement. 
We implement our MatMoE system atop llama.cpp for both CPU and GPU platforms.
Extensive evaluations across three modern MoE models and challenging benchmarks demonstrate that MatMoE achieves $4.0\% \sim 18.3\%$ higher accuracy than state-of-the-art static mixed-precision method. 
Compared to llama.cpp, MatMoE delivers $1.3\times \sim 3.7\times$ speedup on Raspberry Pi 5, and $1.2\times \sim 1.3\times$ speedup on NVIDIA RTX 4090 and Jetson AGX Orin.

\end{abstract}

\maketitle

\section{Introduction}
\label{sec:introduction}

Mixture-of-Experts (MoE) architecture~\cite{dai2024deepseekmoe,liu2024deepseekv2,liu2024deepseekv3,team2024qwen2,yang2025qwen3} has been widely 
adopted in Large Language Models (LLMs).
By dynamically activating only a small subset of experts for each token, MoE models achieve better scalability and performance than dense LLMs with comparable active parameter counts. 
However, their large model size incurs substantial memory overhead, hindering deployment on resource-constrained edge devices.

\begin{figure}[!tb]
\centering
\includegraphics[width=0.9\linewidth]{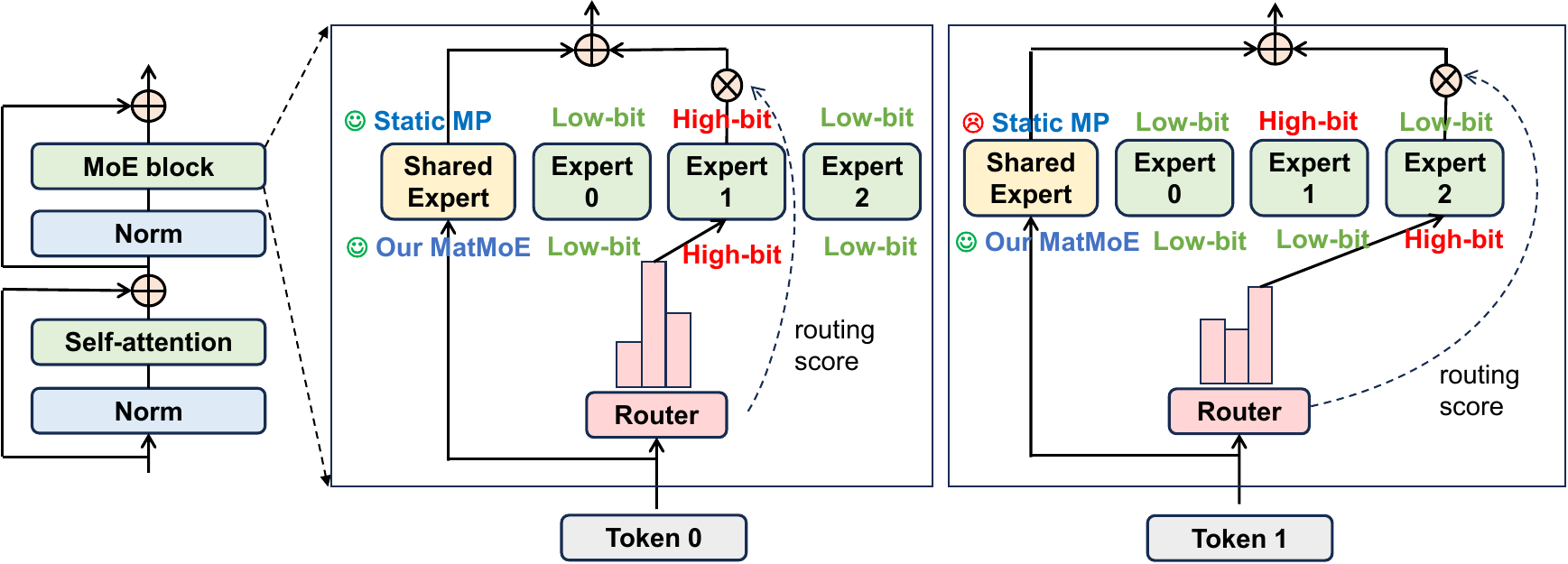}
\caption{MoE architecture and the comparison between the static and our proposed dynamic mixed-precision quantization.
Static method assigns each expert a fixed bit-width offline, regardless of the input token. In contrast, our method dynamically assigns higher precision to more important activated experts and lower precision to less important ones.} 
\vspace{-10pt}
\label{fig:intro}
\end{figure}

Quantization is an effective way to reduce the model size.
Compared to uniform bit-width quantization, mixed-precision quantization~\cite{dong2019hawq,dong2020hawq,yao2021hawq} can achieve better performance under the same memory budget by selectively assigning higher bit-widths to the more critical layers.
Recent studies have focused on mixed-precision quantization on MoE models~\cite{yi2023edgemoe,hu2025moequant,huang2025milo,fu2025eaquant, li2024quantmoe,huangmixture}.
They observe that experts are not equally important and assign different bit-widths to experts according to their importance. 
However, these methods are static, i.e., determining the bit-width allocation strategy based on a calibration dataset and fixing it throughout the decoding steps.
Yet, we observe that the expert importance varies across decoding steps.
As shown in Fig.~\ref{fig:intro}, the static mixed-precision quantization cannot adapt to the dynamic importance of experts, resulting in suboptimal performance.
How to dynamically allocate different bit-widths according to the varying importance of experts remains an important open-ended question.

Furthermore, static quantization methods inherently fail to accommodate the dynamic latency constraints of edge devices, where fluctuating system utilization dictates variable performance targets~\cite{kwon2025dp}.
Restricted to a fixed bit-width allocation, these static methods lack runtime adaptability.
While multi-scale quantization~\cite{park2024any,nair2025matryoshka,kwon2025dp} has emerged as an effective solution for runtime model adaptation by overlaying multiple bit-width variants within a single model, existing works are tailored for dense LLMs and exhibit three critical limitations.
First, they adjust the precision at a coarse model-level~\cite{park2024any} or layer-level~\cite{kwon2025dp,nair2025matryoshka},
which is suboptimal for MoE models.
Second, they either lack an automated mechanism~\cite{park2024any, nair2025matryoshka} or require an expensive fine-tuning process~\cite{kwon2025dp} to determine the precision.
Third, they introduce severe hardware inefficiencies.
Any-Precision LLM~\cite{park2024any} and DP-LLM~\cite{kwon2025dp} are codebook-based non-uniform quantization methods, incurring significant table-lookup overhead for dequantization during inference.
Although Matryoshka Quantization~\cite{nair2025matryoshka} leverages a uniform quantization method,
it requires an extra bit for rounding,
thereby resulting in redundant memory access.

To tackle these challenges,
we propose Matryoshka Mixture-of-Experts (MatMoE),
a dynamic mixed-precision quantization framework for efficient MoE inference featuring algorithm-system co-design.
First, given a stringent memory budget,
we propose the Matryoshka MoE PTQ algorithm,
to derive a parent MoE model, which 
enables experts to operate flexibly at multiple precisions.
Second, to address the inefficiency in memory access,
we propose the floor-and-compensation quantization method
that eliminates the need for an extra rounding bit.
Third, during inference, 
we propose a dynamic bit-width allocation strategy that adaptively assigns different bit-widths to experts at runtime to satisfy the dynamic latency constraint.
Finally, we implement our MatMoE system on top of llama.cpp~\cite{llama_cpp} for both CPU and GPU platforms.
Extensive experimental results demonstrate that 
MatMoE achieves a superior accuracy-efficiency trade-off compared to the state-of-the-art (SOTA) static mixed-precision quantization method.
Our key contributions are as follows:
\begin{itemize}
    \item We observe that expert importance is dynamic across decoding steps in MoE models and propose a Matryoshka
    MoE PTQ algorithm to construct a parent model that inherently supports multi-precision expert execution.

    \item We propose the floor-and-compensation quantization method to eliminate the redundant memory access in the Matryoshka representation.
    
    \item We propose a dynamic bit-width allocation strategy that determines the optimal precisions to individual experts at runtime during inference.
    
    \item We implement an efficient MoE inference system for both CPU and GPU platforms.
     Compared to the llama.cpp baseline, MatMoE delivers $1.3\times \sim 3.7\times$ end-to-end speedup on Raspberry Pi 5, and $1.2\times \sim 1.3\times$ speedup on NVIDIA RTX 4090 and Jetson AGX Orin.
     
    \item We evaluate MatMoE on three modern MoE models across three challenging generative benchmarks.
    MatMoE achieves $4.0\% \sim 18.3\%$ average accuracy improvement compared to the SOTA static mixed-precision quantization method.

\end{itemize}

\section{Background}
\label{sec:background}

\subsection{Mixture-of-Experts}

Mixture-of-Experts (MoE)~\cite{fedus2022switch,jiang2024mixtral,dai2024deepseekmoe,liu2024deepseekv2,liu2024deepseekv3,team2024qwen2,yang2025qwen3} is widely adopted in modern LLMs. As shown in Fig.~\ref{fig:intro}, an MoE block replaces a single feed-forward network with multiple experts and a router that activates the top-$k$ experts for each token. Recent models, such as DeepSeekMoE~\cite{dai2024deepseekmoe,liu2024deepseekv2,liu2024deepseekv3} and QwenMoE~\cite{team2024qwen2,yang2025qwen3}, employ fine-grained routed experts for specialized knowledge and always-active shared experts for common knowledge. The layer output combines the shared-expert outputs with the routed-expert outputs weighted by their routing scores.

\subsection{Quantization for LLM}
Post-training quantization (PTQ) reduces the computational and storage costs by converting weights or activations from FP16 to low-bit integer or floating-point representations.
For dense LLMs, GPTQ~\cite{frantar2022gptq} quantizes weights using a Hessian-based error compensation mechanism.
OmniQuant~\cite{shao2023omniquant} introduces learnable weight clipping and learnable equivalent transformation, achieving better performance.
For MoE models, MoEQuant~\cite{hu2025moequant} incorporates routing scores into GPTQ.
EdgeMoE~\cite{yi2023edgemoe}, QuantMoE-Bench~\cite{li2024quantmoe} and PMQ~\cite{huangmixture} further extend MoE quantization to mixed-precision quantization, achieving better performance.
The SOTA method PMQ~\cite{huangmixture} first assigns different bit-widths to different experts based on their activation frequency and routing score, and then quantizes weights using GPTQ.
However, these quantization methods are static, which determine the bit-width allocation based on a calibration dataset
and fix it through decoding steps,
thus leading to suboptimal performance.

\subsection{Multi-Scale Quantization}

Multi-scale quantization has recently emerged as a promising paradigm to address varying runtime constraints~\cite{kwon2025dp}.
It maintains a single quantized model capable of supporting inference at multiple precisions, thereby offering flexible accuracy-efficiency trade-offs. 
Any-Precision DNN~\cite{yu2021any} first proposes this concept in convolutional neural networks.
Any-Precision LLM~\cite{park2024any} introduces multi-scale quantization to LLM, featuring a codebook-based non-uniform quantization and a specialized GPU software engine.
DP-LLM~\cite{kwon2025dp} further improves Any-Precision LLM by dynamically assigning different bit-widths to layers based on their relative quantization error.
However, the codebook-based non-uniform quantization incurs significant table-lookup overhead for weight dequantization.
Matryoshka Quantization~\cite{nair2025matryoshka} proposes a uniform multi-scale quantization technique.
However, it requires an extra bit for rounding, diminishing the memory bandwidth saving.
Moreover, all these methods ignore the dynamic importance of experts in MoE.
A comparison of these methods with our MatMoE is presented in Table~\ref{tab:related_work_compare}.

\begin{table}[!tb]
\centering
\caption{Comparison of existing multi-scale quantization works with our proposed method.}
\label{tab:related_work_compare}
\resizebox{0.45\textwidth}{!}{
\begin{tabular}{c|cccc}
\toprule
                        & Granularity & Quant. scheme                      & Efficiency \\ \midrule
Any-Precision LLM\cite{park2024any}       & Model  & Non-uniform (codebook)           & Low        \\
DP-LLM\cite{kwon2025dp}                  & Layer  & Non-uniform (codebook)           & Low        \\
Matryoshka Quant.\cite{nair2025matryoshka} & Layer  & Uniform (extra bit rounding)      & Low        \\
\textbf{Ours}                    & Expert & Uniform (floor-and-compensation) & High       \\ \bottomrule
\end{tabular}
}
\end{table}
\section{Motivation}
\label{sec:motivation}

  



\begin{figure*}[!tb]
  \centering
  
  \subfloat[Qwen: Variations of expert ranking changes across decoding steps.]{
      \label{subfig:qwen_expert_ranking}
      \includegraphics[width=0.45\textwidth]{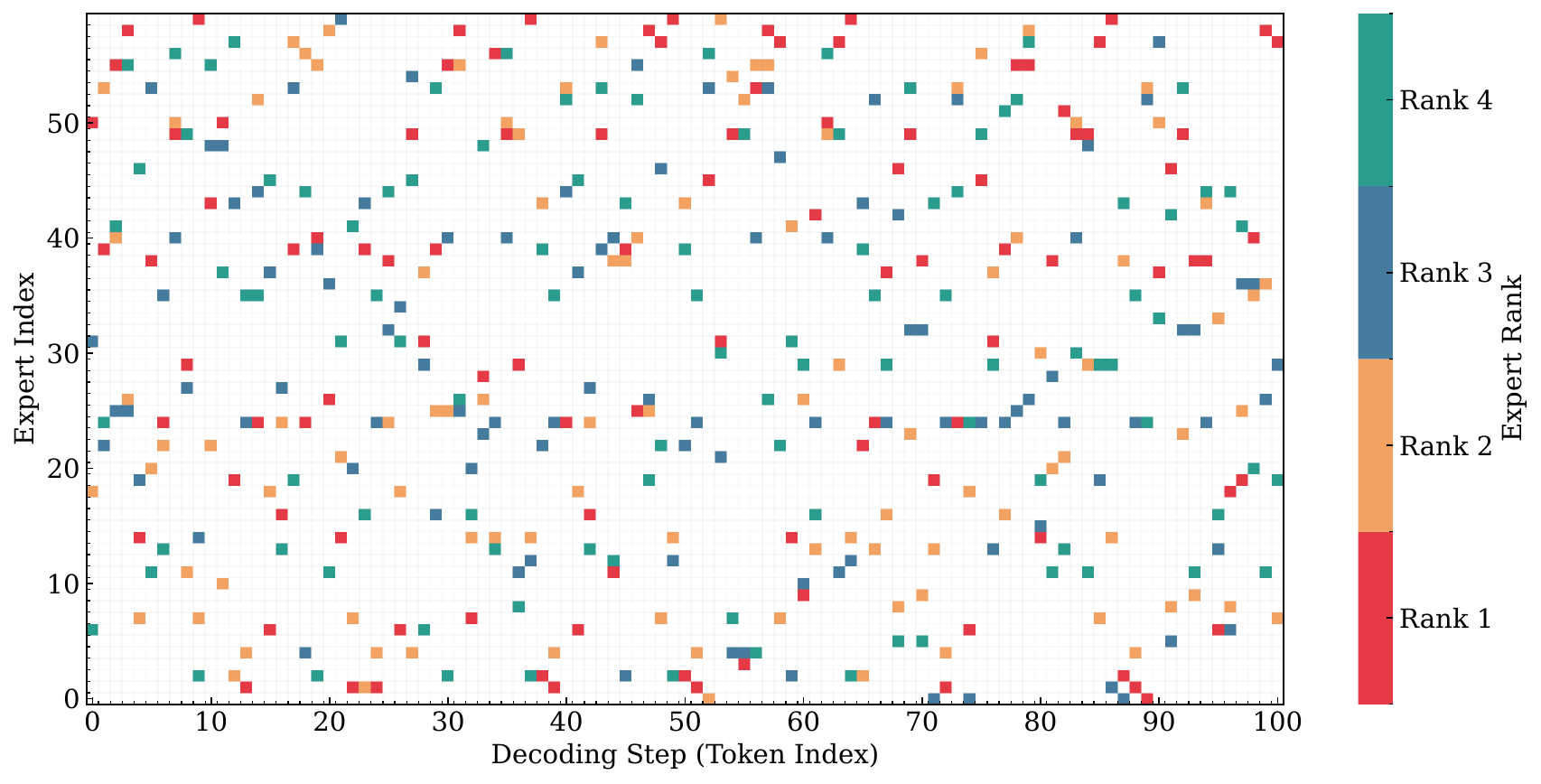}
    }
  \hfill
  \subfloat[Qwen : Variation of routing scores for activated experts.]{
      \label{subfig:qwen_routing_score}
      \includegraphics[width=0.45\textwidth]{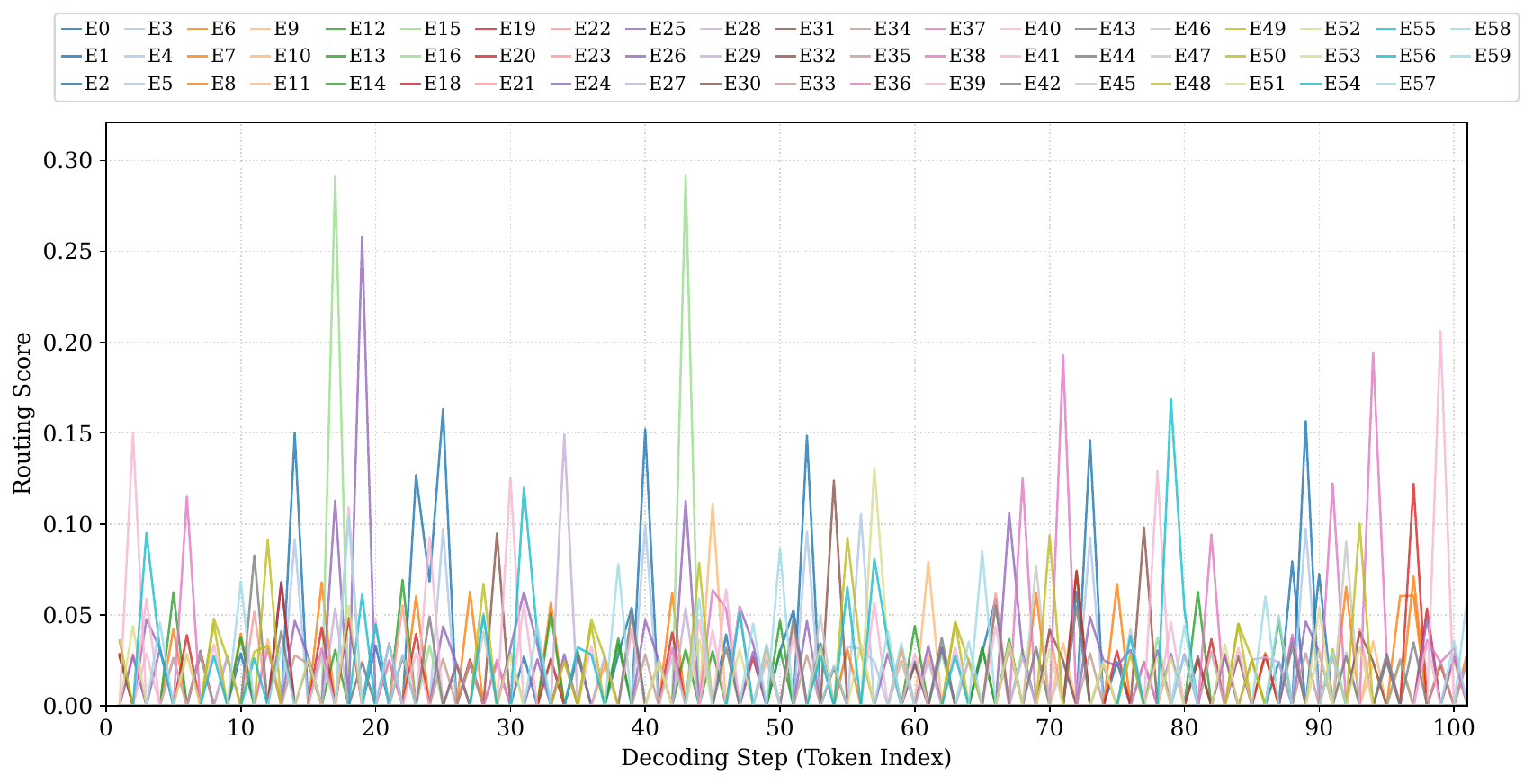}
    }
    
  \caption{
    The routing scores and expert ranking variations in a single layer of Qwen1.5-MoE-A2.7B~\cite{qwen1.5-moe-a2.7b-hf} for the first sample in the GSM8K-CoT benchmark~\cite{cobbe2021training}.
  }
  \label{fig:combined_moe_variation}
\end{figure*}

\textbf{Motivation 1: Dynamic expert importance across decoding steps.}
Existing mixed-precision quantization methods for MoE models treat expert importance—whether measured by routing scores or activation frequencies~\cite{yi2023edgemoe,li2024quantmoe,huangmixture,duanmu2025mxmoe}—as a static metric, determining it offline using a calibration dataset.
However, we find that the expert importance is inherently dynamic, changing significantly across decoding steps.
This is because of the specialization of different experts and the varying semantics of input tokens.
Notably, the load balancing loss used in MoE training is to guarantee sufficient and balanced parameter optimization across the training corpus, rather than to homogenize expert functionality.
Instead, it facilitates the development of highly specialized experts tailored to distinct knowledge domains~\cite{mu2025comprehensive,yang2025qwen3}.

As illustrated in Fig.~\ref{subfig:qwen_expert_ranking},
expert ranking for the same expert varies significantly
across decoding steps even within a single sample.
The dramatic fluctuations of the routing scores in Fig.~\ref{subfig:qwen_routing_score} further 
demonstrate the highly dynamic nature of expert importance.
Consequently, this motivates a dynamic mixed-precision quantization method
that adaptively assigns the optimal bit-width 
based on the expert importance at run time.

\textbf{Motivation 2: Stringent memory constraint and dynamic latency constraint on edge devices.} 
Edge devices have limited memory capacity and dynamic runtime latency requirements since the system utilization fluctuates rapidly~\cite{kwon2025dp}.
Multi-scale quantization has been proposed as an effective solution.
However, existing works are designed for dense LLMs, which exhibit the following limitations.
First, they adjust the precision at the model-level~\cite{park2024any} or layer-level~\cite{kwon2025dp,nair2025matryoshka},
which incurs suboptimal performance for MoE models.
Second, they either lack an automated mechanism for determining the precision~\cite{park2024any, nair2025matryoshka} or require a costly fine-tuning process~\cite{kwon2025dp}.
Consequently, this motivates us to develop a multi-scale quantization method tailored for MoE models with an efficient precision determination mechanism.


\textbf{Motivation 3: Redundant memory access in the Matryoshka representation.}
Although Matryoshka Quantization~\cite{nair2025matryoshka} adopts a uniform quantization scheme to avoid the significant table-lookup overhead in Any-Precision LLM~\cite{park2024any} and DP-LLM~\cite{kwon2025dp},
it has a fundamental limitation that relies on an auxiliary rounding bit, which inadvertently leads to redundant memory accesses.
For example, deriving a 2-bit quantized weight from a 4-bit weight requires fetching the 3rd bit to correctly resolve the rounding logic.
Consequently, 2-bit inference requires 3-bit memory access, degrading the theoretical speedup from $2\times$ down to $1.33\times$. 
Worse still, extracting 3 bits forces a full 4-bit memory access, completely neutralizing any potential throughput improvement.
On the other hand, naively truncating the first 2 bits will lead to significant accuracy degradation. 
This dilemma highlights the urgent need for a novel quantization paradigm that secures accurate inference without the redundant memory access of the extra rounding bit.

\section{Method}
\label{sec:method}

\subsection{Overview}
\label{subsec: overview}

\begin{figure}[!tb]
\centering
\includegraphics[width=0.9\linewidth]{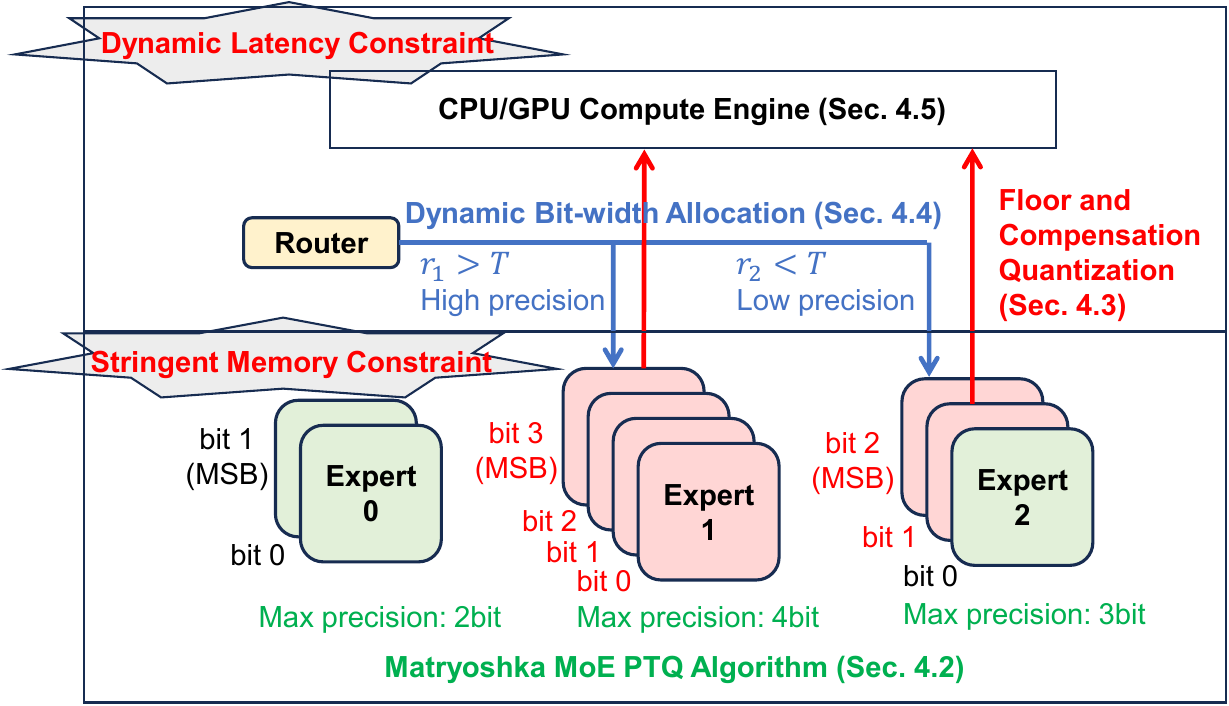}
\caption{Overview of MatMoE.} 
\vspace{-15pt}
\label{fig:overview}
\end{figure}

In this paper, we propose MatMoE,
a dynamic mixed-precision quantization framework for efficient MoE inference featuring algorithm-system co-design.
As shown in Fig.~\ref{fig:overview},
MatMoE consists of a Matryoshka MoE PTQ algorithm that enables experts to operate flexibly at multiple precisions within a strict memory budget (Sec.~\ref{subsec: Matryoshka MoE Training Algorithm}), a floor-and-compensation quantization method addressing the redundant memory access in the Matryoshka representation (Sec.~\ref{subsec: Floor and Compensation Quantization}), and a
dynamic bit-width allocation strategy that determines the optimal precision for individual experts based on their routing scores at run time (Sec.~\ref{subsec:Dynamic Bit-Width Allocation}).
Furthermore, we design efficient MoE compute engines on both CPU and GPU (Sec.~\ref{subsec:System Implementation}).

\subsection{Matryoshka MoE PTQ Algorithm}
\label{subsec: Matryoshka MoE Training Algorithm}

In order to enable experts to perform well at different precisions under the strict memory budget of edge devices, we propose the Matryoshka MoE PTQ algorithm, which consists of two stages.

\textbf{Stage 1: Determine the maximum allowable precision for each expert.} 
Given a strict memory budget on edge devices, we first allocate the maximum allowable precision to each expert, aiming to minimize quantization loss while satisfying the memory constraint.
In contrast to PMQ~\cite{huangmixture}, which relies on
a heuristic metric to guide bit-width allocation, we propose a more principled metric derived directly from the MoE layer's output.
Specifically, the quantization loss of the MoE layer can be approximated using a second-order Taylor expansion as follows:
\begin{equation}
\label{eq:loss approx}
L = \sum_{i=1}^{N} r_i \cdot H_i \left[W_i - Q(W_i)\right]^2
\end{equation}
where $N$ is the number of activated experts,
$W_i$ is the weight of the $i$-th expert,
$Q(\cdot)$ is the quantization function,
$r_i$ and $H_i$ are the routing score and the Hessian value
of the $i$-th expert, respectively.
Given the prohibitive memory and computational overhead of computing the exact Hessian matrix, we approximate it using the Fisher Information Matrix following~\cite{kim2023squeezellm} as below:
\begin{equation}
\label{eq:hessian}
H = \frac{1}{|D|} \sum_{d \in D} g_w g_w^T
\end{equation}
where g represents the gradient of weights and D is the calibration dataset.
Therefore, our proposed metric for each expert, i.e., $r_i \cdot H_i \left[W_i - Q(W_i)\right]^2$, can be calculated offline based on a calibration dataset.
Based on this metric,
we determine the maximum precision allocation leveraging Integer Linear Programming (ILP) optimization, which only takes a few seconds to finish.
The ILP problem is formulated as:


\begin{equation}
\begin{aligned}
& \underset{z_{i, b}}{\arg\min} \; L = \sum_{i=1}^{N} \sum_{b=1}^{B} z_{i, b} \cdot r_i \cdot H_i \left[W_i - Q(W_i)\right]^2 \\
& \text{s.t.} \quad z_{i,b} \in \{0,1\}, \quad \sum_{b=1}^{B} z_{i,b} = 1, \quad \sum_{i=1}^{N} \sum_{b=1}^{B} z_{i,b} \cdot b  \leq P_{\text{budget}}
\end{aligned}
\end{equation}
where $z_{i,b}$ is a binary variable indicating whether the $i$-th expert is quantized to the $b$-th bit-width.
$B$ is the optional bit-width set, which we set as \{2,3,4\}-bit.
$P_{\text{budget}}$ is the equivalent bit precision dictated by the memory budget.

\textbf{Stage 2: PTQ for Matryoshka MoE.}
In this stage, we aim to obtain a single MoE model that performs well at multiple precisions.
Specifically, we leverage OmniQuant~\cite{shao2023omniquant} to quantize weights and train the quantization parameters in a layer-wise manner, considering all candidate precisions to enable experts to adapt to different bit-widths.
The objective function we adopt is defined as:
\begin{equation}
\label{eq:matryoshka training loss}
\underset{\gamma, \beta}{\arg\min} \; L = \sum_{p\in P} \lambda_p \cdot \mathcal{L} \left[ F(Q_p(W,\gamma,\beta ),X), F(W, X) \right]
\end{equation}
where $P$ is the bit-width set including the maximum precision $p_h$ and the lowest precision $p_l$ of each expert.
We set $p_l$ to 2-bit and $p_h$ is derived from Stage 1.
$\lambda_p$ is the hyperparameter.
$\mathcal{L}$ is the mean squared error loss.
$W$ and $X$ are the FP16 weights and activations, respectively.
$F(\cdot)$ is the forward function.
$Q_p(\cdot)$ is the quantization function quantizing weights to $p$-bit,
$\gamma$ and $\beta$ are the learnable quantization parameters shared across different bit-widths.

\subsection{Floor-and-Compensation Quantization}
\label{subsec: Floor and Compensation Quantization}
Within the Matryoshka MoE PTQ algorithm, the quantization function plays a pivotal role, as it directly dictates the performance of the low-precision expert sharing weights with its high-precision counterpart.
Different from the original Matryoshka quantization method~\cite{nair2025matryoshka},
we propose the floor-and-compensation quantization for the low precision to address the redundant memory access.

We first illustrate the quantization function for the maximum precision $p_h$.
We adopt the standard OmniQuant~\cite{shao2023omniquant} scheme, which is defined as follows:
\begin{equation}
\begin{aligned}
&W_{p_h} = \operatorname{clamp}\left( \operatorname{round} \left( \frac{W}{s} \right) + z,\; 0,\; 2^{p_h} - 1 \right), \\
\; &s = \frac{\gamma \max(W) - \beta \min(W)}{2^{p_h} - 1}, \;  z = -\operatorname{round} \left(\frac{\beta \min(W)}{s}\right)
\end{aligned}
\end{equation}
where $W$ and $W_{p_h}$ denote the original FP16 and $p_h$-bit quantized weights, respectively.
$s$ is the scaling factor.
$z$ is the zero point.
$\gamma$ and $\beta$ are two learnable quantization parameters.

\begin{figure}[!tb]
\centering
\begin{minipage}{0.4\linewidth}
    \centering
    \captionof{table}{Expectation shift of directly truncating the first ${p_l}$ bits.}
    \resizebox{0.9\linewidth}{!}{
\begin{tabular}{c|c|c}
\toprule
$W_{p_h}$  & $W_{p_l}$ (round) & $W_{p_l}$ (floor) \\ \midrule
0,1   & 0            & 0            \\
2,3   & 4            & 0            \\
4,5   & 4            & 4            \\
6,7   & 8            & 4            \\
8,9   & 8            & 8            \\
10,11 & 12           & 8            \\
12,13 & 12           & 12           \\
14,15 & 12           & 12           \\ \midrule
7.5   & 7.5          & 6.0           \\ \bottomrule
\end{tabular}
}
\label{tab:round_floor_value}
\end{minipage}%
\hfill
\begin{minipage}{0.5\linewidth}
    \centering
    \includegraphics[width=\linewidth]{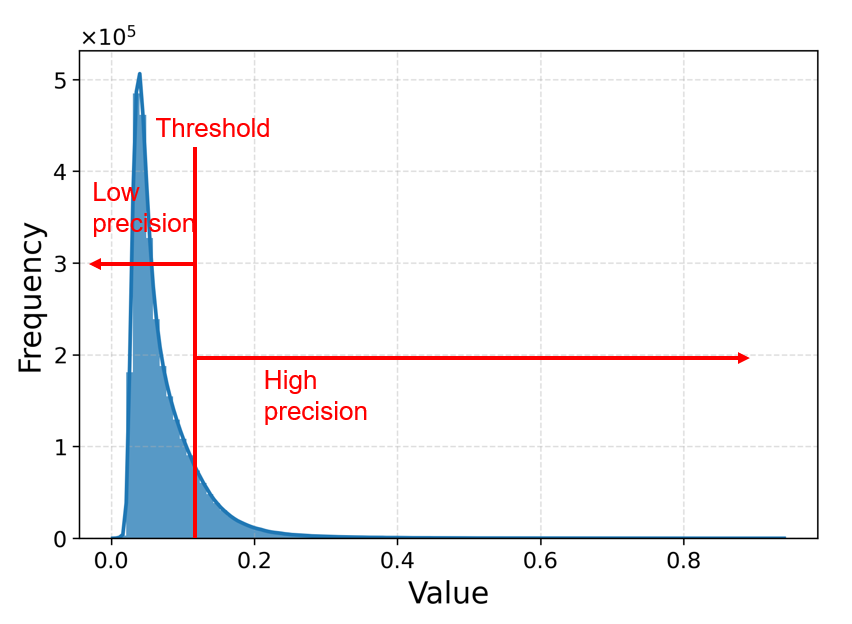}
    \caption{Routing score distribution.}
    \label{fig:routing_score_distribution}
\end{minipage}
\end{figure}

For the lower precision $p_l$, we aim to extract the first $p_l$ bits from the $p_h$-bit weight.
However, the Matryoshka quantization~\cite{nair2025matryoshka} requires an extra bit, i.e., the $(p_l + 1)$-th bit, for rounding, which is:
\begin{equation}
W_{p_l} = \operatorname{clamp}\left( \operatorname{round} \left( \frac{W_{p_h}}{2^{p_h-p_l}} \right), \; 0,\; 2^{p_l} - 1 \right) \times 2^{p_h-p_l}
\end{equation}
This quantization scheme significantly diminishes the memory bandwidth saving.
On the other hand, naively truncating the first $p_l$ bits as in Eq.~\ref{eq:floor}, results in significant accuracy degradation.
For example, as shown in Table~\ref{tab:Comparison of different quantization schemes}, 
directly truncating the first 2 bits from the original 4-bit weight
leads to a performance collapse, reducing accuracy to 0\% on all three benchmarks. 
We find that this degradation is caused by the weight expectation shift.
As shown in Table~\ref{tab:round_floor_value},
there is an expectation shift of 1.5 between
directly truncating the first 2 bits and the original 4-bit weight.
\begin{equation}
\label{eq:floor}
W_{p_l} = \operatorname{clamp}\left( \operatorname{floor} \left( \frac{W_{p_h}}{2^{p_h-p_l}} \right), \; 0,\; 2^{p_l} - 1 \right) \times 2^{p_h-p_l}
\end{equation}

To this end, we propose the floor-and-compensation quantization method.
The key insight is that we use a compensation term to compensate the expectation shift caused by the truncation operation.
The mathematical derivation is as follows.

\textbf{Mathematical Derivation of the Compensation Term.}
Considering truncating the first $p_l$ bits of the $p_h$-bit weights and let $\Delta = p_h - p_l$ denote the number of truncated bits, the original high-precision weight can be perfectly reconstructed if we keep the remainder as below:
\begin{equation}
W_{p_h} = \operatorname{floor} \left( \frac{W_{p_h}}{2^{\Delta}} \right) \times 2^\Delta + R
\end{equation}
where $R$ represents the truncated residual (the discarded lower $\Delta$ bits), meaning $R \in \{0, 1, \dots, 2^\Delta - 1\}$.
Naively truncating the first $p_l$ bits ignores $R$, leading to an expectation shift of the weight values.

\begin{figure}[!tb]%
  \centering
  \hspace*{-3pt}
    \subfloat[]{
        \label{subfig:weight_distribution_and_quant_level}
        \includegraphics[width=0.24\textwidth]{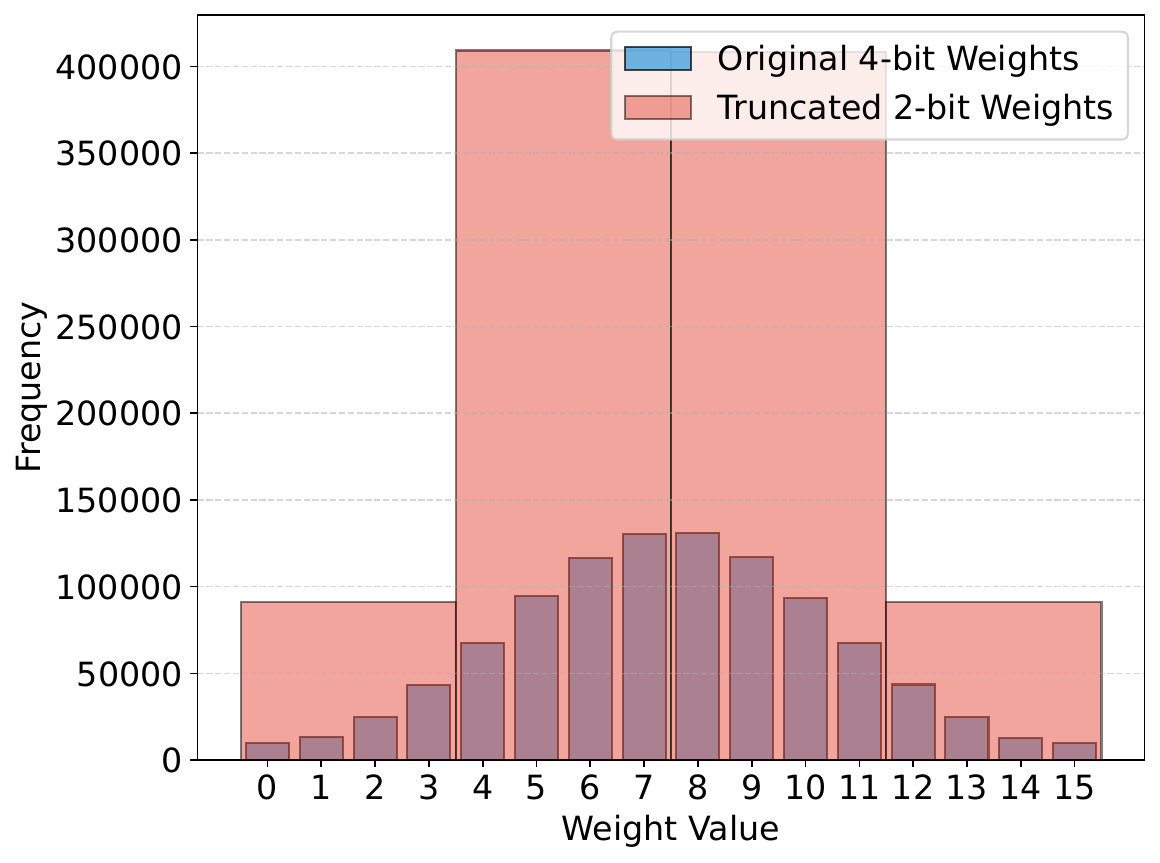}
      }
      \subfloat[]{
        \label{subfig:weight_quant_error}
        \includegraphics[width=0.23\textwidth]{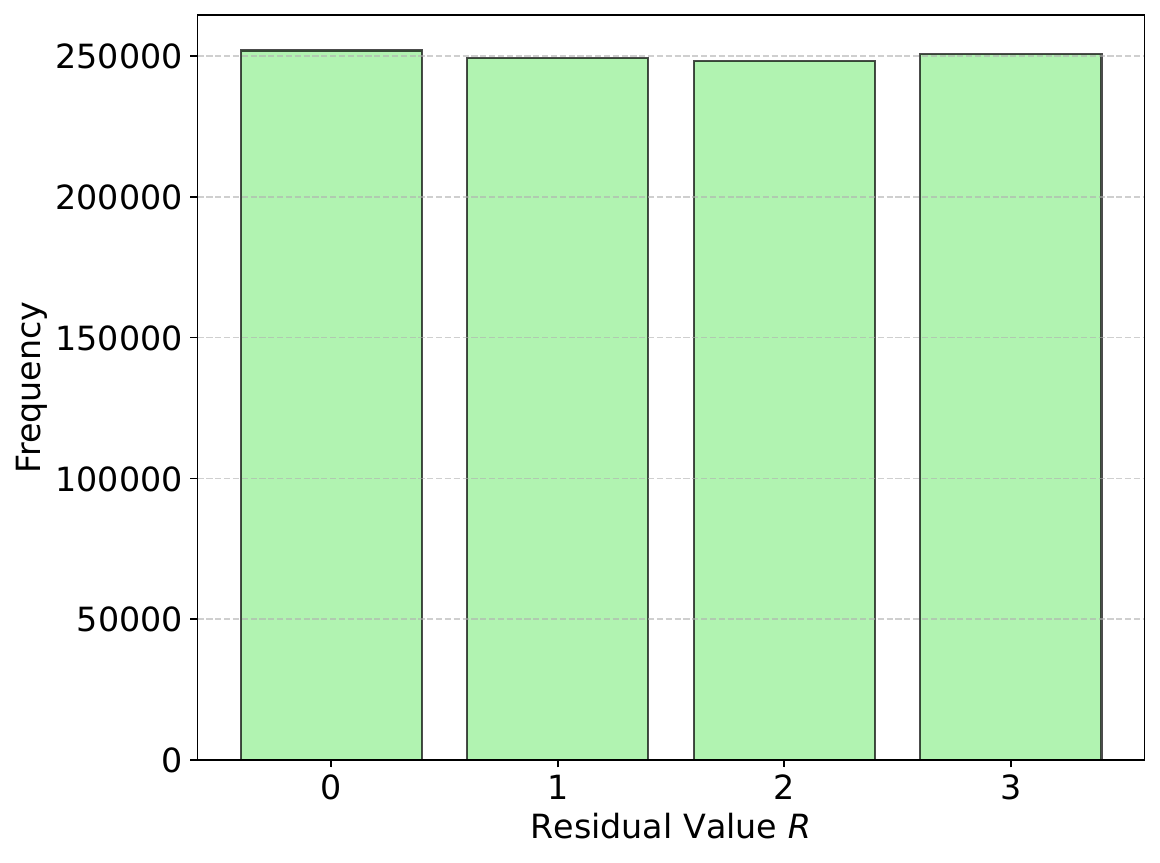}
      }
      
  \caption{
(a) Distribution of the model weights quantized to 4-bit and the truncated 2-bit  weights.
(b) Distribution of the residual between the 4-bit weights and the 2-bit weights.
}
  \label{fig:residual distribution}
\end{figure}

For the original model weights which follow the normal distribution,
the residual $R$ follows a uniform distribution as shown in Fig.~\ref{fig:residual distribution}. 
Thus, the expectation of the residual $R$ is:
\begin{equation}
E[R] = \frac{1}{2^\Delta} \sum_{k=0}^{2^\Delta - 1} k = \frac{1}{2^\Delta} \frac{(2^\Delta - 1)2^\Delta}{2} = \frac{2^\Delta - 1}{2} = \frac{2^{p_h - p_l} - 1}{2}
\end{equation}

To ensure that our truncated low-precision weights remain statistically unbiased, we add this expected error back as a constant compensation term $C=E[R]$.
By introducing $C$, we compensate for the expectation shift caused by the floor operation, resulting in our final floor-and-compensation function:
\begin{equation}
\label{eq:floor and compensation}
W_{p_l} = \operatorname{clamp}\left( \operatorname{floor} \left( \frac{W_{p_h}}{2^{p_h-p_l}} \right), \; 0,\; 2^{p_l} - 1 \right) \times 2^{p_h-p_l} + C
\end{equation}
which is applied in both training and inference.
Thanks to our floor-and-compensation quantization,
we achieve accuracy comparable to, or even surpassing, the original Matryoshka Quantization~\cite{nair2025matryoshka} as shown in Table~\ref{tab:Comparison of different quantization schemes} without requiring any extra rounding bit.
Moreover, introducing $C$ does not increase the memory bandwidth overhead, and it only incurs negligible computational cost.

\begin{table}[!tb]
\centering
\caption{Accuracy impact of different quantization schemes.}
\label{tab:Comparison of different quantization schemes}
\scalebox{0.9}{ 
\begin{tabular}{c|c|c|c}
\toprule
                       & GSM8K-CoT & HumanEval & MBPP \\ \midrule
Matryoshka Quant.~\cite{nair2025matryoshka}             & 21.9      & 14.6      & 17.0 \\ \midrule
Floor                  & 0.0       & 0.0       & 0.0  \\ \midrule
Floor-and-Compensation & 24.5      & 15.2      & 15.8 \\ \bottomrule
\end{tabular}
}
\end{table}

\subsection{Dynamic Bit-Width Allocation}
\label{subsec:Dynamic Bit-Width Allocation}

To adapt to the dynamic expert importance across the decoding steps,
we propose our dynamic bit-width allocation strategy.
Prior studies have established that the routing score serves as a highly robust proxy for an expert's contextual importance~\cite{hu2025moequant,huang2025milo,huangmixture}. Leveraging this insight, we propose a lightweight threshold-based mechanism: during inference, activated experts with routing scores exceeding a pre-determined threshold are executed in high precision, while the remaining experts are executed in low precision.
By adjusting the threshold, we can achieve an arbitrary average bit-width, which enables the system to adapt to various latency constraints.


However, determining the optimal threshold to guarantee that the system satisfies strict latency constraints emerges as a critical challenge.
We propose a binary search algorithm to 
search for the optimal threshold for the given latency constraint,
which is reflected by the target precision $P_\text{target}$.
We apply a binary search algorithm over a routing score set $S$,
which is collected on a calibration dataset as shown in Fig.~\ref{fig:routing_score_distribution}
to find the optimal threshold that strictly bounds the expected
average bit-width to $P_\text{target}$.
The detailed procedure is illustrated in Alg.~\ref{alg:binary_search}.
Notably, the search process is highly efficient, executing with only sub-millisecond latency.


\begin{algorithm}[!tb]
\caption{Binary Search for Optimal Threshold}
\label{alg:binary_search}
\begin{algorithmic}[1]
\REQUIRE Routing score set $S$, Target precision $P_{target}$, High precision $p_h$, Low precision $p_l$, Tolerance $\epsilon$, Max iterations $N_{max}$
\STATE $T_{low} \leftarrow 0$, $T_{high} \leftarrow \max(S)$
\FOR{$iter = 1$ \TO $N_{max}$}
    \STATE $T_{mid} \leftarrow (T_{low} + T_{high}) / 2$
    \STATE $N_{high} \leftarrow \sum_{s \in S} \mathbb{I}(s \ge T_{mid}), N_{low} \leftarrow |S| - N_{high}$
    \STATE $P_{actual} \leftarrow (N_{high} \cdot p_h + N_{low} \cdot p_l) / |S|$
    
    \IF{$|P_{actual} - P_{target}| < \epsilon$}
        \STATE \textbf{break}
    \ELSIF{$P_{actual} > P_{target}$}
        \STATE $T_{low} \leftarrow T_{mid}$
    \ELSE
        \STATE $T_{high} \leftarrow T_{mid}$ 
    \ENDIF
\ENDFOR
\RETURN $T_{mid}$
\end{algorithmic}
\end{algorithm}

\subsection{System Implementation}
\label{subsec:System Implementation}

\begin{figure}[!tb]
\centering
\includegraphics[width=0.95\linewidth]{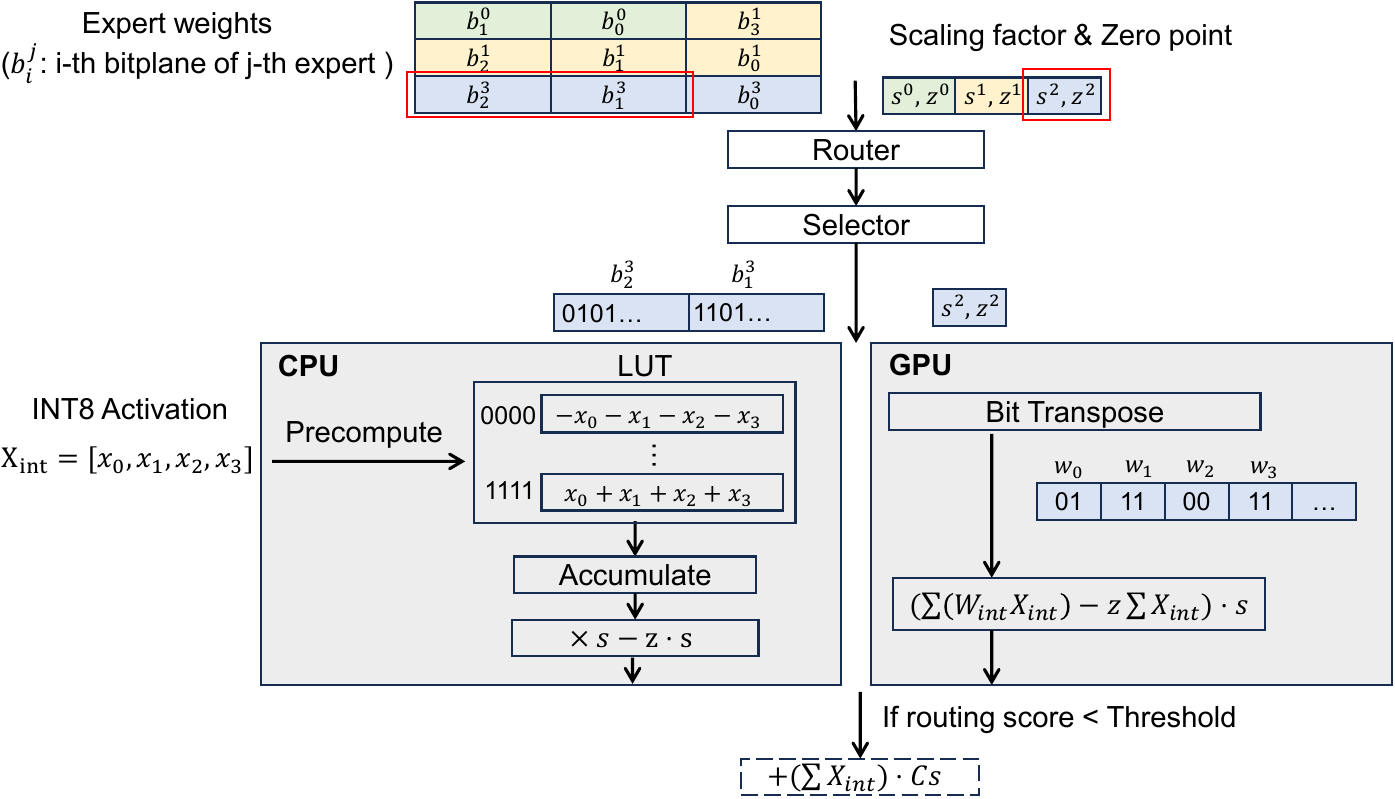}
\caption{System implementation of MatMoE.} 
\vspace{-10pt}
\label{fig:System implementation of MatMoE}
\end{figure}

We design our MatMoE system on both CPU and GPU based on llama.cpp~\cite{llama_cpp},
which is the most widely used LLM inference engine for edge devices. 
The overall system implementation of MatMoE is illustrated in Fig.~\ref{fig:System implementation of MatMoE}.
To facilitate efficient memory access for variable bit-widths, we adopt a bitplane-based representation~\cite{park2024any,wei2025t} for the expert weights.
During inference, the execution precision of each expert is dynamically dictated by whether its routing score surpasses a pre-determined threshold (Sec.~\ref{subsec:Dynamic Bit-Width Allocation}).
At the core of our compute engine, we develop highly optimized GEMV kernels tailored for both CPU and GPU platforms, as detailed below.
After the main computation,
we apply a lightweight compensation term to the 
output of the low-precision experts (Sec.~\ref{subsec: Floor and Compensation Quantization}).

On CPU, we develop a LUT-based compute engine.
The LUT-based GEMV kernel is first proposed in T-MAC~\cite{wei2025t} for dense LLM.
During inference, it precomputes all possible products between the activation and all binary patterns, constructing a LUT.
 The binary matrices of weights serve as indices to retrieve the precomputed results from the LUT, which are then accumulated to obtain the final output.
However, it does not support MoE computation.
To overcome this, we modify the kernel for MoE by introducing fine-grained, expert-level address offset control.
Furthermore, we share the LUT between the gate projection and the up projection of different experts to avoid redundant precomputation and storage overhead, since they receive identical input.

On GPU, we develop our efficient GPU compute engine.
Since table lookup operation incurs severe memory bank conflicts on GPU,
we adopt a computation strategy fundamentally distinct from the CPU engine.
As illustrated in Fig.~\ref{fig:System implementation of MatMoE}, our custom GEMV kernel is composed of the following two operations.
 First, we perform an in-register bit-transpose operation to convert the bit-plane weight layout to the standard weight representation using efficient bitwise logic, i.e., bit-wise masking, shifting and OR operation.
The reconstructed weights are cast to INT8 format and densely packed into 32-bit registers, four weights per register.
Then, we execute the matrix multiplication against INT8 activations by leveraging NVIDIA's 4-way integer dot-product (dp4a) instructions.
Furthermore, we apply kernel fusion techniques to integrate the gate projection, up projection, and SiLU non-linear activation into a single kernel, which completely eliminates the latency of writing intermediate activations back to the global memory.

\section{Experiment}
\label{sec:experiment}

\subsection{Experimental Setup}
\label{subsec:Experimental Setup}

\textbf{Models and Benchmarks.}
We evaluate MatMoE on three modern MoE models, i.e, Qwen1.5-MoE-A2.7B (Qwen)~\cite{qwen1.5-moe-a2.7b-hf}, 
DeepSeek-V2-Lite-16B (DeepSeek)~\cite{liu2024deepseekv2},
and OLMoE-1B-7B-Instruct (OLMoE)~\cite{muennighoff2024olmoe}.
Following~\cite{kwon2025dp}, we evaluate the accuracy using lm-evaluation-harness~\cite{eval-harness}
on the challenging benchmarks, which have sufficient token generation lengths.
Specifically, we evaluate on 8-shot GSM8K-CoT~\cite{cobbe2021training},
0-shot HumanEval~\cite{chen2021evaluating} and 3-shot MBPP~\cite{austin2021program}.
GSM8K-CoT aims to evaluate the model's mathematical reasoning capabilities, while HumanEval and MBPP are utilized to measure its performance on code generation and programming tasks.
The metric we report is the exact-match and flexible-extract
accuracy for GSM8K-CoT,
pass@1 accuracy for HumanEval and MBPP.

\textbf{Quantization Configuration.}
We quantize the weights with a group size of 128.
We quantize all linear layers in the MoE model except for the router, which has a negligible parameter count.
We do not quantize the Key-Value projection in the Multi-Head Latent Attention (MLA) of DeepSeek either due to its marginal parameter size.
Furthermore, for the linear layers in the self-attention and shared experts, we apply 4-bit quantization, while for the routed experts, we adopt mixed-precision quantization.
The same quantization configuration is applied for both our method and the baselines.

\textbf{Calibration and Hyperparameters.}
For our Matryoshka MoE PTQ algorithm,
we leverage a calibration dataset consisting of 128 randomly
selected 2048-token segments from C4 dataset~\cite{raffel2020exploring}
and train the quantization parameters for 40 epochs using a batch size of 4.
We utilize AdamW~\cite{loshchilov2019decoupled} optimizer with zero weight decay,
and set the learning rate to 1e-2.
We set $\lambda_p$ in Eq.~\ref{eq:matryoshka training loss} to 1.0.
For evaluation, we use a batch size of 1, considering the edge inference scenario.

\textbf{Baselines.} 
For accuracy evaluation, we use the SOTA mixed-precision quantization method PMQ~\cite{huangmixture} as the baseline, testing under both 3-bit and 4-bit memory budgets.
For the prefill phase,
we use the maximum precision for each expert,
since lower precision weights only provide a marginal speed improvement in this compute-bound scenario.
For latency evaluation,
we compare our MatMoE system with llama.cpp~\cite{llama_cpp},
the most widely used LLM inference engine for edge devices
on both CPU and GPU.
We report the average Time-Per-Output-Token (TPOT).

\textbf{Hardware Platforms.} 
For CPU evaluation, we choose Raspberry Pi 5 with a quad-core ARM Cortex-A76 processor.
For GPU evaluation,
our selected hardware platforms include NVIDIA Jetson AGX Orin and NVIDIA RTX 4090.

\subsection{Accuracy Evaluation}
\label{subsec:Accuracy Evaluation}


\begin{table}[!tb]
\centering
\caption{Accuracy of FP16 models.}
\label{tab:Accuracy of FP16}
\scalebox{0.9}{ 
\begin{tabular}{c|c|c|c|c}
\toprule
                          &      & GSM8K-CoT & HumanEval & MBPP \\ \midrule
Qwen    & FP16  & 62.9      & 32.9      & 39.4 \\
DeepSeek & FP16  & 39.4     & 28.0        & 43.2 \\
OLMoE   & FP16  & 41.0      & 31.1      & 27.8 \\ \bottomrule
\end{tabular}
}
\end{table}

\begin{table}[!tb]
\centering
\caption{Accuracy comparison under 3-bit memory budget.}
\label{tab:Accuracy comparison under 3-bit memory budget}
\scalebox{0.7}{ 
\begin{tabular}{c|c|ccc|ccc|ccc}
\toprule
                          & Dataset          & \multicolumn{3}{c|}{GSM8K-CoT}                & \multicolumn{3}{c|}{HumanEval}                & \multicolumn{3}{c}{MBPP}                      \\ \cmidrule{2-11} 
                          & Target Prec. & 2.25          & 2.50          & 2.75          & 2.25          & 2.50          & 2.75          & 2.25          & 2.50          & 2.75          \\ \midrule
\multirow{2}{*}{Qwen}     & PMQ              & 32.2          & 41.6          & 48.8          & 17.7          & 17.1          & 23.8          & 20.0          & 23.2          & 22.0          \\
                          & MatMoE           & \textbf{50.6} & \textbf{51.5} & \textbf{52.4} & \textbf{26.8} & \textbf{28.7} & \textbf{28.7} & \textbf{30.8} & \textbf{30.8} & \textbf{31.0} \\ \midrule
\multirow{2}{*}{DeepSeek} & PMQ              & 18.4          & 25.9          & \textbf{31.1} & 10.4          & 12.8          & 17.7          & 25.2          & 24.4          & \textbf{27.6} \\
                          & MatMoE           & \textbf{29.8} & \textbf{29.6} & \textbf{31.1} & \textbf{16.5} & \textbf{18.3} & \textbf{18.3} & \textbf{25.6} & \textbf{25.4} & 26.2          \\ \midrule
\multirow{2}{*}{OLMoE}    & PMQ              & 9.9           & 15.6          & \textbf{22.8}          & 3.7           & 3.0           & 3.7           & 1.4           & 1.6           & 2.6           \\
                          & MatMoE           & \textbf{11.0}              &  \textbf{18.7}             &   22.7       & \textbf{12.8}   &   \textbf{14.6} & \textbf{14.6} &    \textbf{11.8}           &     \textbf{12.0}     &   \textbf{12.2}   \\ \bottomrule
\end{tabular}
}
\vspace{-5pt}
\end{table}

\begin{table*}[!tb]
\centering
\caption{Accuracy comparison under 4-bit memory budget.}
\label{tab:Accuracy comparison under 4-bit memory budget}
\scalebox{0.78}{ 
\begin{tabular}{l|l|ccccccc|ccccccc|ccccccc}
\toprule
 & \multicolumn{1}{c|}{Dataset}
 & \multicolumn{7}{c|}{GSM8K-CoT}
 & \multicolumn{7}{c|}{HumanEval}
 & \multicolumn{7}{c}{MBPP} \\
\cmidrule(lr){2-2}\cmidrule(lr){3-9}\cmidrule(lr){10-16}\cmidrule(lr){17-23}
 & \multicolumn{1}{c|}{Target Prec.}
& 2.25 & 2.50 & 2.75 & 3.00 & 3.25 & 3.50 & 3.75 
& 2.25 & 2.50 & 2.75 & 3.00 & 3.25 & 3.50 & 3.75
& 2.25 & 2.50 & 2.75 & 3.00 & 3.25 & 3.50 & 3.75 \\
\midrule

\multirow{2}{*}{Qwen}
& PMQ    
& 32.2 & 41.6 & 48.8 & 56.0 & 52.8 & 56.0 & \textbf{57.3}
& 17.7 & 17.1 & 23.8 & 20.1 & 18.9 & 25.0 & 30.5
& 20.0 & 23.2 & 22.0 & 27.0 & 30.8 & 32.6 & 34.8 \\

& MatMoE 
& \textbf{50.6} & \textbf{52.1} & \textbf{55.3} & \textbf{55.7} & \textbf{56.5} & \textbf{56.9} & 56.9
& \textbf{31.1} & \textbf{34.2} & \textbf{34.2} & \textbf{33.5} & \textbf{33.5} & \textbf{33.5} & \textbf{35.4}
& \textbf{31.4} & \textbf{34.2} & \textbf{36.0} & \textbf{36.2} & \textbf{36.4} & \textbf{36.4} & \textbf{36.6} \\
\midrule

\multirow{2}{*}{DeepSeek}
& PMQ    
& 18.4 & 25.9 & 31.2 & 34.7 & 34.8 & \textbf{38.9} & \textbf{39.3}
& 10.4 & 12.8 & 17.7 & 22.6 & 18.9 & 25.0 & \textbf{25.6}
& 25.2 & 24.4 & 27.6 & 30.6 & 33.4 & 34.4 & 36.4 \\

& MatMoE 
& \textbf{32.1} & \textbf{36.2} & \textbf{36.2} & \textbf{36.6} & \textbf{36.6} & 36.5 & 36.8
& \textbf{25.0} & \textbf{26.2} & \textbf{26.8} & \textbf{26.2} & \textbf{26.6} & \textbf{26.6} & \textbf{25.6}
& \textbf{39.6} & \textbf{41.6} & \textbf{42.6} & \textbf{41.0} & \textbf{42.2} & \textbf{41.8} & \textbf{42.4} \\
\midrule

\multirow{2}{*}{OLMoE}
& PMQ    
& 9.9 & 15.6 & 22.8 & 27.7 & 23.5 & 28.4 & 31.5
& 3.7 & 3.0 & 3.7 & 3.7 & 6.1 & 6.7 & 9.8
& 1.4 & 1.6 & 2.6 & 6.8 & 6.4 & 5.2 & 13.4 \\

& MatMoE 
& \textbf{29.1} & \textbf{31.5} & \textbf{34.8} & \textbf{36.5} & \textbf{37.1} & \textbf{38.9} & \textbf{39.5}
& \textbf{24.4} & \textbf{25.0} & \textbf{28.1} & \textbf{26.8} & \textbf{26.8} & \textbf{29.9} & \textbf{28.1}
& \textbf{23.4} & \textbf{25.6} & \textbf{26.6} & \textbf{26.2} & \textbf{26.8} & \textbf{26.8} & \textbf{26.8} \\
\bottomrule

\end{tabular}
}

\end{table*}

\begin{figure*}[!tb]
    \centering
    \scriptsize
    \begin{tabular}{cccc}
        \textcolor{color1}{\rule{1em}{1em}} llama.cpp & 
        \textcolor{color2}{\rule{1em}{1em}} MatMoE & \\
    \end{tabular} \\[0.5em]
\centerline{\includegraphics[width=\linewidth]{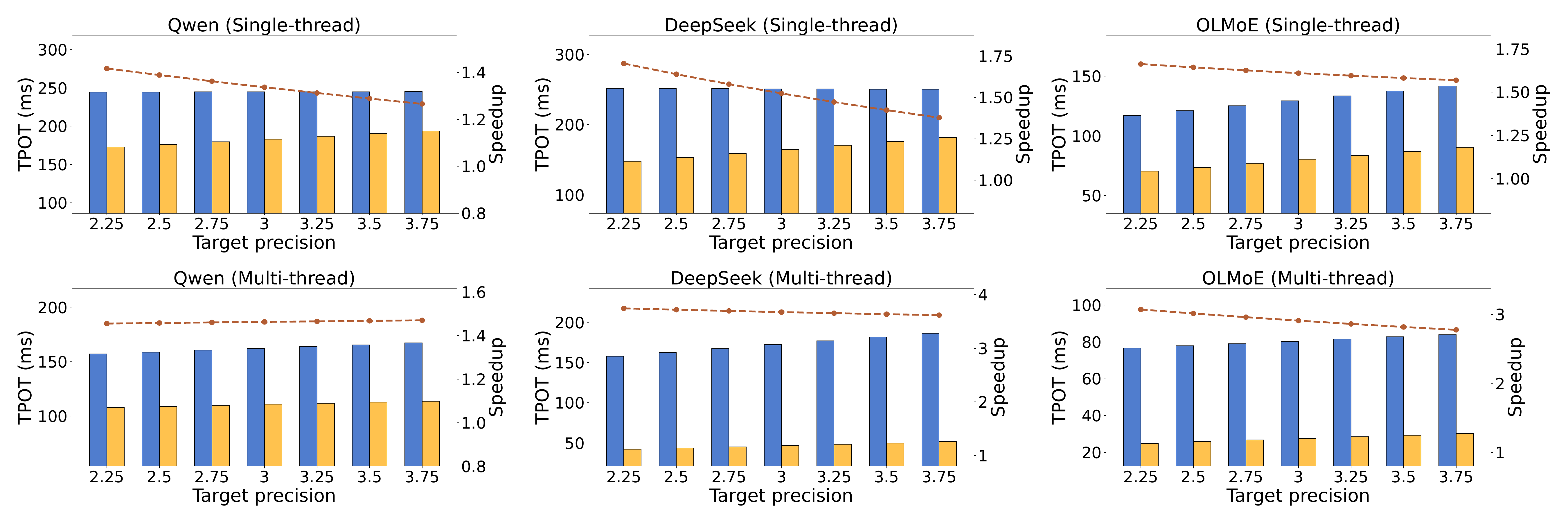}}
\caption{Latency comparison on Raspberry Pi 5 across three models at different target precisions.}
\label{fig:latency_cpu}
\vspace{-5pt}
\end{figure*}

We first present the accuracy of the FP16 model on three benchmarks in Table~\ref{tab:Accuracy of FP16}.
We demonstrate the accuracy of PMQ and our MatMoE under 3-bit and 4-bit memory budgets in Table~\ref{tab:Accuracy comparison under 3-bit memory budget} and Table~\ref{tab:Accuracy comparison under 4-bit memory budget}, respectively. 
MatMoE consistently exhibits superior performance over the SOTA method PMQ.
Specifically, under 3-bit memory budget, MatMoE achieves 4.0\% to 9.8\% average accuracy improvement across three models.
Under 4-bit memory budget, MatMoE
achieves  7.5\% to 18.3\%  average accuracy improvement across three models.
The substantial accuracy improvement is attributed to MatMoE’s capability to dynamically allocate bit-widths to experts based on their importance.
It is also worth noting that PMQ necessitates storing multiple models for different target precisions, whereas our method natively supports these diverse target precisions using only a single model.

\subsection{Latency Evaluation}
\label{subsec:System Performance}


\begin{figure*}[!tb]
    \centering
    \scriptsize
    \begin{tabular}{cccc}
        \textcolor{color1}{\rule{1em}{1em}} llama.cpp & 
        \textcolor{color2}{\rule{1em}{1em}} MatMoE & \\
    \end{tabular} \\[0.5em]
\centerline{\includegraphics[width=\linewidth]{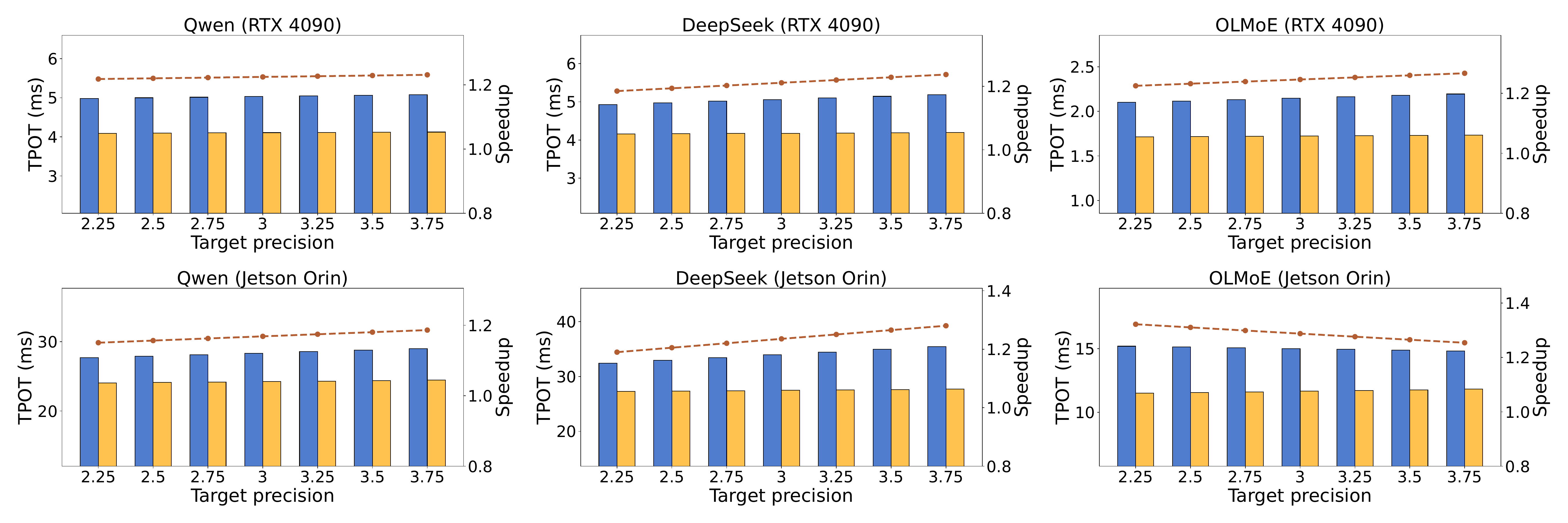}}
\caption{Latency comparison on NVIDIA RTX 4090 and Jetson Orin across three  models at different target precisions.}
\label{fig:latency_gpu}
\end{figure*}

\begin{figure}[!tb]
\centering
\begin{minipage}{0.46\linewidth}
\centering
\includegraphics[width=\linewidth]{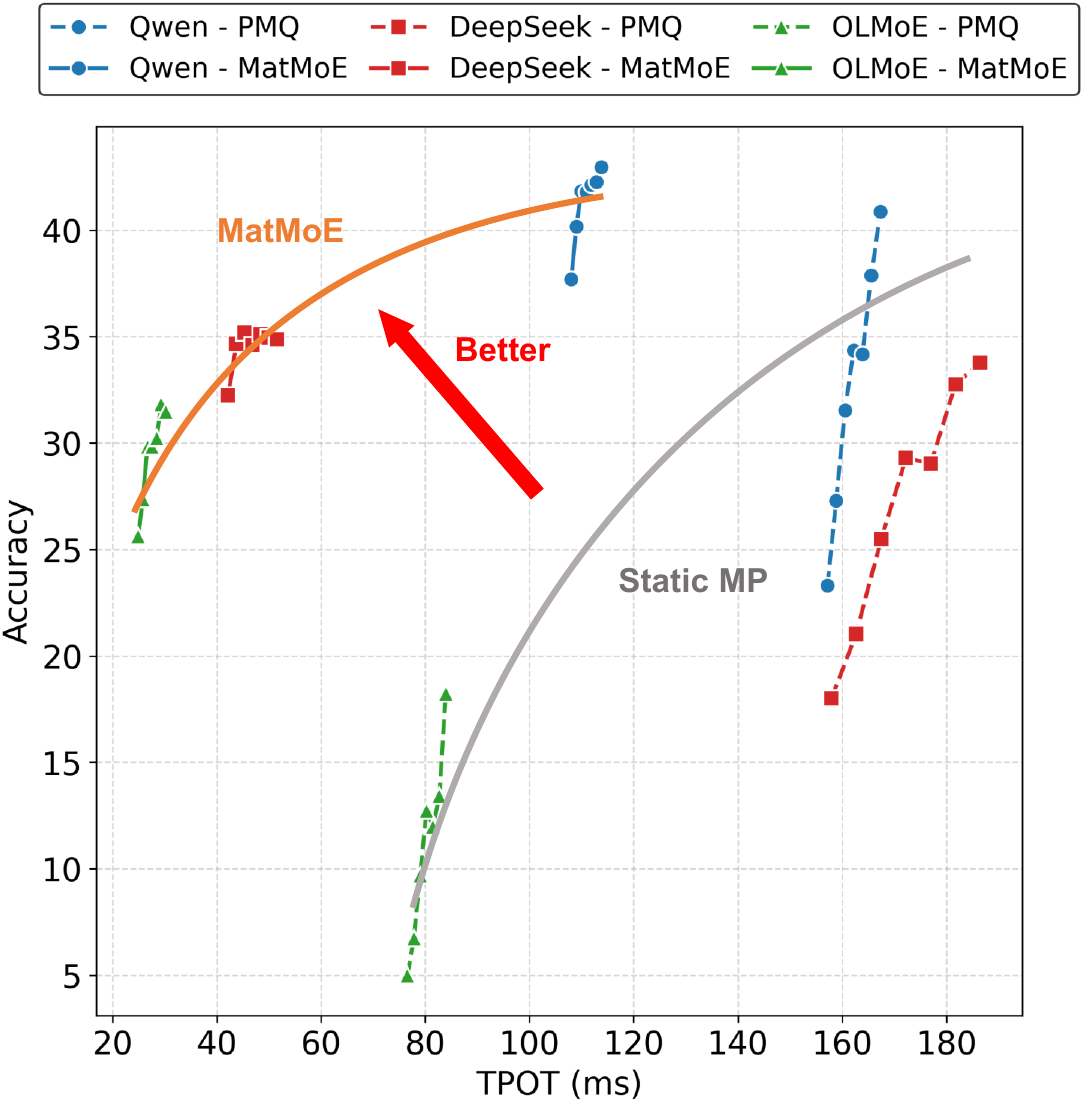}
\caption{Accuracy-efficiency trade-off of MatMoE and the SOTA static mixed-precision quantization method.} 
\label{fig:Accuracy-efficiency trade-off}
\end{minipage}
\hfill
\begin{minipage}{0.46\linewidth}
    \centering
    \includegraphics[width=\linewidth]{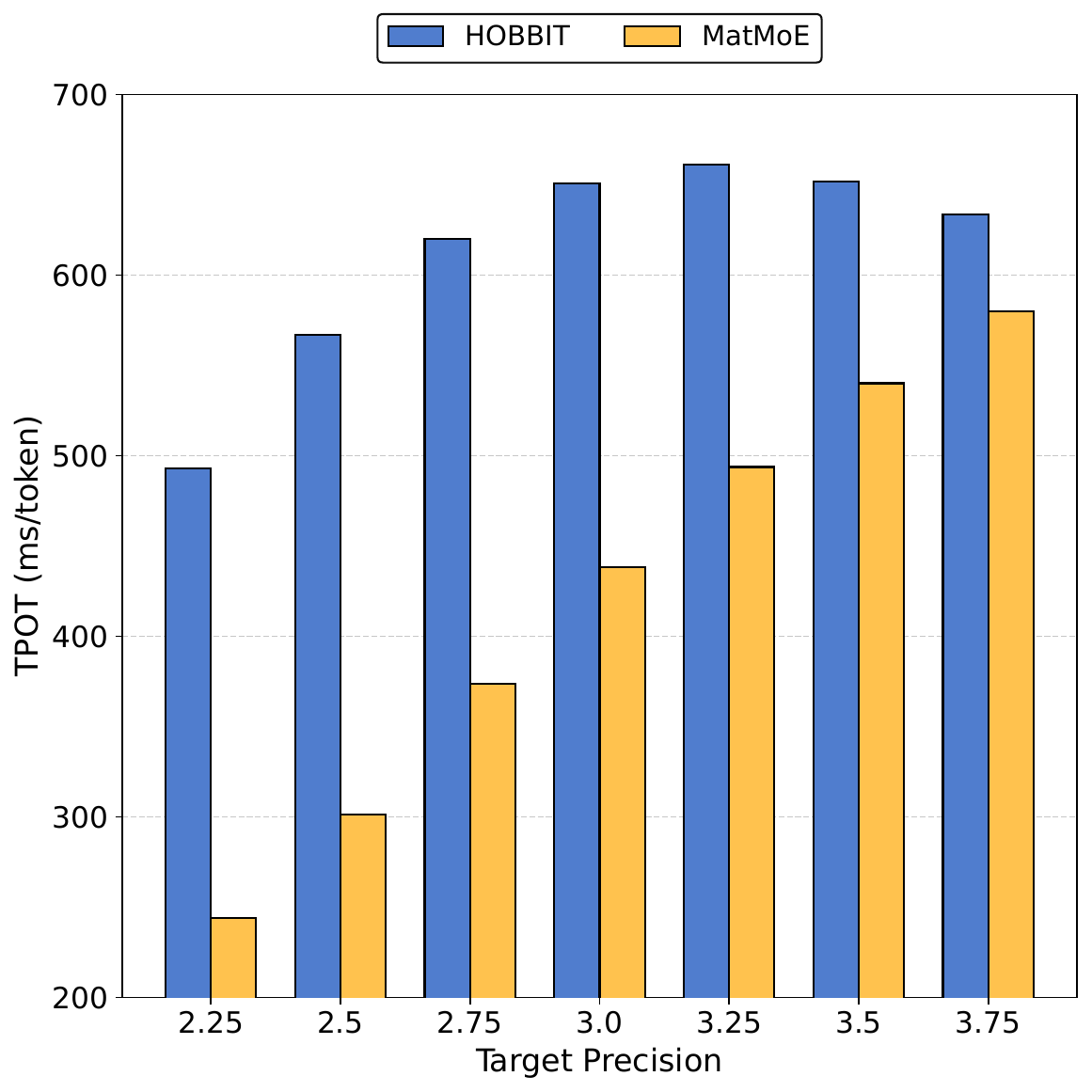}
    \caption{Latency comparison of HOBBIT~\cite{tang2024hobbit} and MatMoE under Qwen3-30B-A3B offloading on Raspberry Pi 5.}
    \label{fig:matmoe_vs_hobbit}
\end{minipage}
\vspace{-5pt}
\end{figure}

As illustrated in Fig.~\ref{fig:latency_cpu},
MatMoE achieves consistent end-to-end speedup over llama.cpp on the CPU across various target precisions.
Under single-thread setting, MatMoE achieves average speedup of 1.3$\times$,
1.5$\times$, 1.6$\times$ for Qwen, DeepSeek, and OLMoE, respectively.
Under multi-thread setting,
MatMoE achieves 1.5$\times$, 3.7$\times$, and 2.9$\times$ speedup across three models, respectively.
Qwen exhibits a smaller speedup due to the larger proportion of shared experts, for which we use 4-bit quantization.
For the latency evaluation on GPU,
as shown in Fig.~\ref{fig:latency_gpu},
MatMoE demonstrates $1.2\times \sim 1.3\times$ speedup over llama.cpp
on both NVIDIA RTX 4090 and Jetson Orin.
The relatively flat TPOT curves on GPU are due to limitations of our current kernel implementation, where low-bit weights are first converted to INT8 before dp4a execution, making different precisions have similar runtime cost. 
Furthermore, we show the accuracy-efficiency trade-off between the average accuracy and the multi-thread TPOT in Fig.~\ref{fig:Accuracy-efficiency trade-off}.
MatMoE achieves a superior accuracy-efficiency trade-off compared to the SOTA static mixed-precision quantization baseline.


\subsection{Ablation Study}
\label{subsec:Ablation Study}

\textbf{Effects of individual components.}
To evaluate the contribution of each component, we compare four configurations in Table~\ref{tab:ablation}.

\emph{Optimized low-bit kernels.}
We first replace the original llama.cpp backend of PMQ with our optimized low-bit kernels while retaining its static mixed-precision allocation. 
PMQ + Our Kernels reduces the average TPOT from 80.3 to 27.4~ms/token, corresponding to a $2.93\times$ speedup over PMQ + llama.cpp. Meanwhile, the two configurations achieve identical accuracy. These results demonstrate that our optimized kernels substantially improve execution efficiency without affecting the numerical behavior.

\emph{Dynamic bit-width allocation.}
Under the same optimized backend, PMQ + Our Kernels and our MatMoE achieve similar TPOTs. 
However, MatMoE improves the average accuracy from 22.8\% to 35.3\%, demonstrating the effectiveness of MatMoE's dynamic allocation strategy, which adapts expert bit-widths to token-level routing importance during decoding.

\emph{Floor-and-compensation.}
To evaluate the contribution of floor-and-compensation, we construct Matryoshka + Dynamic, which combines the original Matryoshka quantization scheme with our dynamic bit-width allocation. 
Unlike MatMoE, this baseline requires accessing an additional bitplane for rounding. 
Compared with this baseline, MatMoE reduces the average TPOT from 30.5 to 27.6~ms/token, while slightly improving the average accuracy from 34.4\% to 35.3\%. 
This confirms that F\&C eliminates redundant rounding-bit accesses without compromising model accuracy.

\begin{table}[!tb]
\centering
\caption{Latency and accuracy comparison under 4-bit memory budget. We evaluate OLMoE on GSM8k-CoT on Raspberry Pi 5 with 4-thread inference.}
\label{tab:ablation}
\scalebox{0.65}{
\begin{tabular}{c|c|ccccccc}
\toprule
                                                                                                      & Target precision & 2.25 & 2.50 & 2.75 & 3.00 & 3.25 & 3.50 & 3.75 \\ \midrule
\multirow{2}{*}{PMQ + Llama.cpp}                                                                      & TPOT         & 76.6 & 77.8 & 79.1 & 80.3 & 81.5 & 82.7 & 83.9 \\
                                                                                                      & Accuracy              & 9.9  & 15.6 & 22.8 & 27.7 & 23.5 & 28.4 & 31.5 \\ \midrule
\multirow{2}{*}{PMQ + Our kernels}                                                                    & TPOT          & 24.9 & 25.6 & 26.6 & 27.3 & 28.4 & 29.2 & 30.0 \\
                                                                                                      & Accuracy              & 9.9  & 15.6 & 22.8 & 27.7 & 23.5 & 28.4 & 31.5 \\ \midrule
\multirow{2}{*}{Matryoshka~\cite{nair2025matryoshka} + Dynamic}                                                                 & TPOT          & 30.1 & 30.3 & 30.4 & 30.5 & 30.7 & 30.8 & 31.0 \\
                                                                                                      & Accuracy              & 26.1 & 31.1 & 34.9 & 35.6 & 37.0 & 37.5 & 38.9 \\ \midrule
\multirow{2}{*}{\begin{tabular}[c]{@{}c@{}} Our MatMoE   \\      (Matryoshka + F\&C +Dynamic)\end{tabular}} & TPOT          & 25.0 & 25.8 & 26.7 & 27.6 & 28.5 & 29.4 & 30.2 \\
                                                                                                      & Accuracy              & 29.1 & 31.5 & 34.8 & 36.5 & 37.1 & 38.9 & 39.5 \\ \bottomrule
\end{tabular}
}
\vspace{-10pt}
\end{table}

\textbf{Offloading efficiency.}
When a model's parameter size exceeds the available DRAM capacity, a portion of its weights must be offloaded to SSD storage. 
MatMoE is also effective in this regime because its nested bitplane representation reduces the amount of expert data transferred between SSD and DRAM. 
To evaluate its offloading efficiency, we conduct experiments on Qwen3-30B-A3B~\cite{yang2025qwen3} using a least-recently-used (LRU) cache replacement policy. 
The cache capacity is set to accommodate all non-expert weights and 25\% of the experts at 4-bit precision.
We compare MatMoE with HOBBIT~\cite{tang2024hobbit}, a state-of-the-art expert offloading method that stores multiple precision experts and dynamically loads high- or low-precision experts according to their importance. 
Since HOBBIT does not provide its source code, we implement a simulator to evaluate the end-to-end latency.
As shown in Fig.~\ref{fig:matmoe_vs_hobbit}, MatMoE achieves substantial end-to-end speedups over HOBBIT under offloading. 
Across different target precisions, HOBBIT requires $493.06\sim 661.23 $~ms/token on Raspberry Pi 5, whereas MatMoE requires only $243.78\sim579.71$~ms/token, corresponding to a $1.1\times \sim 2.0\times $ speedup. 
The speedup stems from the reduced SSD-to-DRAM traffic when the required precision of an expert changes across tokens. 
When an expert switches from 2-bit to 4-bit precision, HOBBIT must load a separate 4-bit copy from SSD, whereas MatMoE loads only the two additional bitplanes. 
Conversely, when the required precision changes from 4-bit to 2-bit, HOBBIT must load a separate 2-bit copy, whereas MatMoE simply ignores the two unnecessary bitplanes without additional I/O. 
Therefore, MatMoE directly reduces offloading traffic and thereby improves the end-to-end latency.
\section{Conclusion}
\label{sec:Conclusion}

In this paper,
we propose MatMoE,
a dynamic mixed-precision quantization framework for efficient MoE inference.
MatMoE enables experts to adaptively operate at the optimal precision based on their dynamic importance through our Matryoshka MoE PTQ algorithm and dynamic bit-width allocation,
thus achieving better performance.
MatMoE also achieves superior inference efficiency
due to our floor-and-compensation quantization and efficient system implementation.
Compared to llama.cpp, MatMoE delivers end-to-end speedups of $1.3\times \sim 3.7\times$ on the CPU platform and $1.2\times \sim 1.3\times$ on the GPU platform.
MatMoE demonstrates $4.0\% \sim 18.3\%$ higher average accuracy across three modern MoE models compared to SOTA static mixed-precision quantization method.

\section{Acknowledgment}
\label{sec:acknowledgment}

This work was supported in part by the Beijing Municipal Science and Technology Program under Grant Z241100004224015, in part by New Generation Artificial Intelligence-National Science and Technology Major Project under Grant 2025ZD0122502, in part by NSFC under Grant 92464104, Grant 62495102, and Grant 62341407, in part by 111 Project under Grant B18001.

\newpage

\bibliographystyle{ACM-Reference-Format}
\bibliography{quantization}

\end{document}